\documentclass[conference]{IEEEtran}
\IEEEoverridecommandlockouts
\usepackage{cite}
\usepackage{amsmath,amssymb,amsfonts}
\usepackage{algorithmic}
\usepackage{graphicx}
\usepackage{textcomp}
\usepackage[table]{xcolor}
\usepackage{booktabs}
\usepackage{multirow}
\usepackage{placeins}
\usepackage{tabularx}
\usepackage{enumitem}
\usepackage{hyperref}
\def\BibTeX{{\rm B\kern-.05em{\sc i\kern-.025em b}\kern-.08em T\kern-.1667em\lower.7ex\hbox{E}\kern-.125emX}}

\begin{document}

\title{CoGeo-GS: Concept-Driven and Geometry-Aware Multi-Object Removal in 3D Scenes}

\author{
\IEEEauthorblockN{
Yuanxiang Ni\textsuperscript{1} \quad
Xianliang Huang\textsuperscript{2,$\dagger$} \quad
Chenhang Ma\textsuperscript{3} \quad
Chen Xiao\textsuperscript{4}
}
\IEEEauthorblockN{
Yuewen Ma\textsuperscript{2} \quad
Ruxin Wang\textsuperscript{5,*} \quad
Hao Zhang\textsuperscript{5,*}
}
\vspace{0.15cm}
\IEEEauthorblockA{
\textsuperscript{1}Southern University of Science and Technology, Shenzhen, China \quad
\textsuperscript{2}PICO, ByteDance, Beijing, China \\
\textsuperscript{3}Zhejiang University, Hangzhou, China \quad
\textsuperscript{4}Fudan University, Shanghai, China \\
\textsuperscript{5}Shenzhen Institutes of Advanced Technology, Chinese Academy of Sciences, Shenzhen, China
\thanks{Emails: 12433317@mail.sustech.edu.cn, huangxl21@m.fudan.edu.cn, \{rx.wang, h.zhang10\}@siat.ac.cn. $^\dagger$ Project leader. $^*$ Corresponding authors.}
}
}

\maketitle

\begin{abstract}
Multi-object removal in 3D scenes is challenging due to severe occlusions, semantic entanglement, and the difficulty of maintaining geometric and multi-view consistency. Existing 3D Gaussian Splatting (3DGS) methods perform well for single-object editing but scale poorly to multi-object scenarios, often requiring repetitive optimization and yielding unstable geometry in removed regions. We propose CoGeo-GS, a concept-driven framework for controllable multi-object removal in 3D scenes. CoGeo-GS assigns concept-aware semantic tags to Gaussians, enabling flexible object selection and reducing interference between foreground objects and background structures within a single optimization stage. To recover plausible geometry, we introduce a geometry-aware completion pipeline that combines monocular depth priors with diffusion-based refinement and boundary-aligned blending. A geometry-regularized refinement strategy further stabilizes reconstruction and preserves multi-view consistency. Experiments demonstrate that CoGeo-GS outperforms existing methods in visual quality and reconstruction fidelity.
\end{abstract}

\begin{IEEEkeywords}
3D Gaussian Splatting, multi-object removal, concept-aware tagging, monocular depth priors, multi-view consistency
\end{IEEEkeywords}

\section{Introduction}
3D scene editing is becoming a core capability for applications such as virtual reality (VR)~\cite{mao2024live}, autonomous driving~\cite{yang2025drivemoe}, and robot interaction~\cite{zhu20243d, huang2026semantic}. Object removal requires plausible completion of the occluded background for controllable 3D editing. Real-world cluttered scenes involve multiple interacting objects and occlusions, requiring precise multi-target specification and geometric completion. Despite recent advances in neural rendering and scene representation, achieving reliable multi-object removal remains challenging. Moreover, maintaining global multi-view consistency and coherent appearance across reconstructed scene further increases the task difficulty.


Recent works based on 3D Gaussian Splatting (3DGS)~\cite{kerbl20233d} have demonstrated promising results for single-object editing~\cite{chen2024gaussianeditor,ye2024gaussian} and inpainting~\cite{wang2024learning}. However, moving from single-object to multi-object scenarios exposes two core issues. First, most approaches still rely on repeated per-object optimization, which becomes computationally prohibitive and prone to error accumulation as the number of targets grows. Second, nearby objects and shared background structures are often entangled in the representation, so editing one object easily perturbs others and leaves occluded regions poorly constrained, leading to semantic interference, distorted geometry, and unstable depth where content has been removed.

To tackle these limitations, we introduce CoGeo-GS, a concept-driven geometric completion framework for high-quality multi-object removal in 3D Gaussian scenes. CoGeo-GS first performs concept-aware semantic tagging on Gaussians. Each primitive receives a compact concept encoding. This allows flexible, user-controllable selection of arbitrary object subsets within a single optimization stage. It also disentangles foreground objects from background structures. As a result, mutual interference is greatly reduced while the editable, spatially explicit 3D representation is preserved.

On top of this concept-level decomposition, we design a geometry-aware completion pipeline that jointly exploits monocular depth priors and generative refinement. We first extract structural priors from a monocular depth foundation model and align them to the 3DGS coordinate system to obtain seed depth in the removed regions. These seeds are then refined by a diffusion-based inpainting module that recovers dense, fine-grained depth for occluded areas, while a boundary-aligned blending scheme enforces smooth transitions and surface continuity between completed and preserved regions. This hybrid strategy yields geometrically stable, high-fidelity reconstructions in the edited areas.
To further stabilize appearance and maintain multi-view coherence, we introduce a geometry-regularized refinement stage. We first freeze geometry for a few steps and then allow only small, trust-region-limited geometric updates while refining appearance under image-level supervision from the edited views. 

In summary, our main contributions are: (1) We propose CoGeo-GS, a concept-driven framework for controllable multi-object removal in 3D Gaussian Splatting, performing object-level editing in a single optimization stage while maintaining an explicit, editable 3D representation. (2) We introduce a hybrid geometric completion pipeline anchoring monocular depth priors to scene geometry and refining missing regions via diffusion-based depth completion, enabling accurate, consistent reconstruction without boundary artifacts. (3) We design a geometry-regularized refinement strategy decoupling appearance adaptation from constrained geometric updates via freezing and trust-region clipping, improving cross-view consistency and fusion stability. (4) Extensive experiments on diverse scenes show CoGeo-GS consistently outperforms existing multi-object removal methods in visual quality and reconstruction fidelity.


\section{Related Work}
\subsection{Object-Centric 2D \& 3D Inpainting}
Image inpainting is a fundamental task in creating appropriate content for the target area. 
Traditional patch-based
approaches~\cite{criminisi2004region,ruvzic2014context} and GAN-driven methods~\cite{zeng2021cr, suvorov2022resolution} are designed to handle regular regions but struggle with complex occlusions. 
Diffusion models~\cite{ho2020denoising,song2020score} demonstrate superior ability in generating semantically coherent content for large missing areas~\cite{lugmayr2022repaint,xie2023smartbrush}.
Building upon the advancements in 2D image inpainting, the extension to 3D scenes reconstructed by NeRF and 3DGS is still a challenging task. 
3D inpainting demands handling intricate spatial structures and multi-view consistency.
While NeRF-based approaches~\cite{mirzaei2023spin,huang2024hi,huang2026nerf} achieve partial success with static volumetric representations, their effectiveness remains constrained by inherent architectural limitations. 
Explicit 3DGS inpainting
 methods like InFusion~\cite{liu2024infusion} and GaussianEditor~\cite{chen2024gaussianeditor} primarily address single-object-centric removal from an existing static Gaussian Splatting scene, overlooking multiple incidental objects present during scene capture. These gaps motivate our concept-driven, scale-aligned geometric completion framework in an explicit 3DGS representation.


\subsection{Multi-Object Removal}
Recent advancements in scene editing~\cite{prabhu2023inpaint3d,huang2023iddr} have investigated various 2D and 3D representations for removing or modifying multiple objects in complex scenes.
In the 2D domain, large diffusion models enable high-fidelity inpainting via generative priors from massive image datasets, but they lack 3D consistency and geometric awareness. To address this, techniques like Inpaint3D~\cite{prabhu2023inpaint3d} use diffusion priors to guide NeRF reconstructions, facilitating object removal and view-consistent scene completion.
Gaussian-based methods such as Gaussian Grouping~\cite{ye2024gaussian} support instance-level segmentation and editing through mask-guided clustering. However, they struggle to remove multiple objects, as optimizing the corresponding Gaussian primitives with consistent and high-quality textures remains difficult.
Our work combines concept-aware tagging with depth-guided completion to support controllable multi-object removal in complex 3DGS scenes.


\section{Method}
\begin{figure*}[t]
  \centering
  \includegraphics[width=\linewidth]{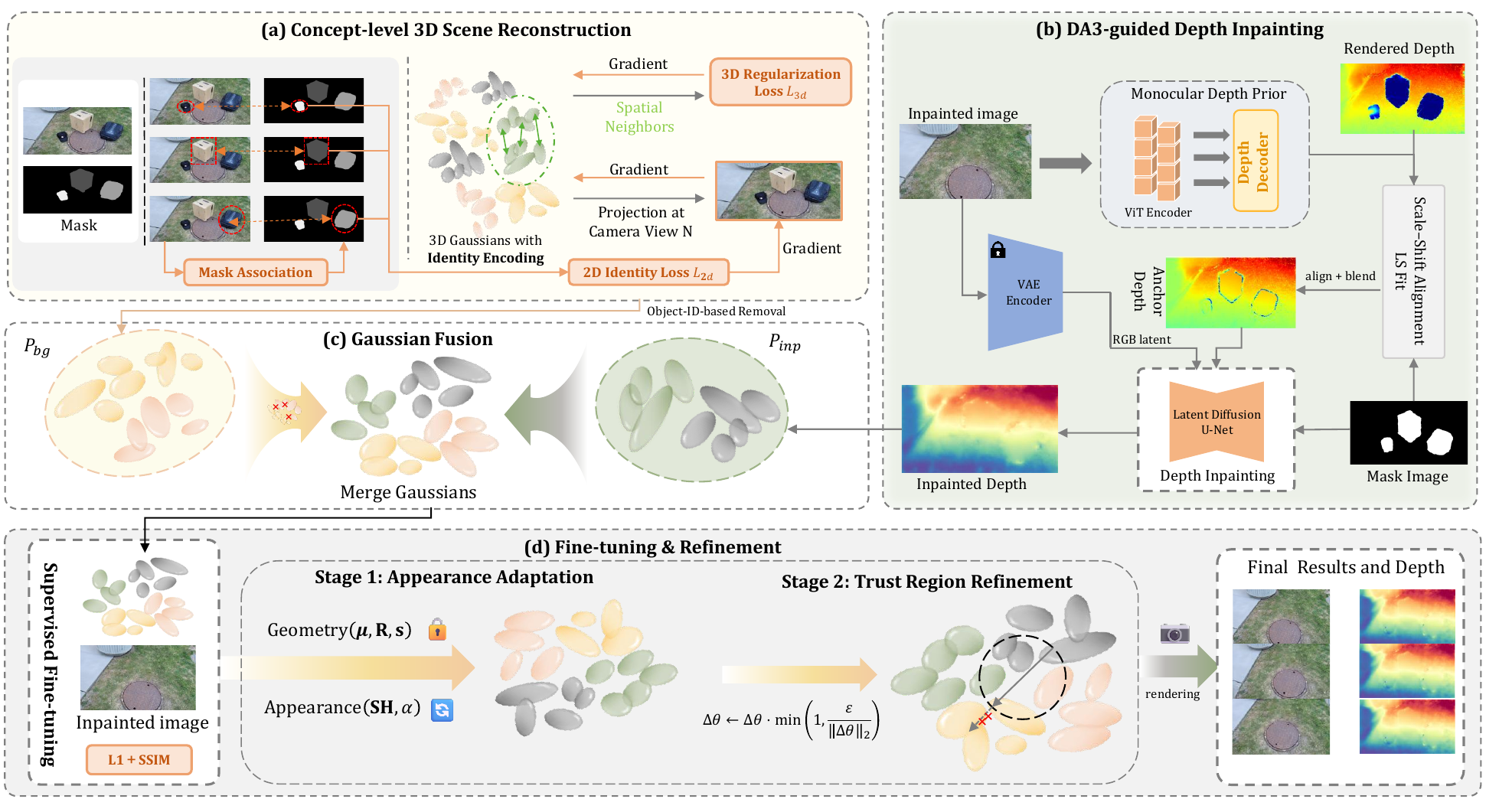}
  \caption{Overview of the CoGeo-GS framework. (a) We first distill text-driven segmentation masks into 3D Gaussians to assign object-level identities. (b) Then, geometric voids are completed by conditioning a latent diffusion model on scale-aligned monocular depth priors (depth anchors) and VAE-encoded RGB latents. (c) The completed geometry is fused with the background via overlap-aware pruning. (d) A two-stage refinement strategy first optimizes appearance, then geometric constraint was subsequently employed to eliminate inconsistencies in fusion seams while preserving structural integrity.  
  }
  \label{fig:pipeline}
\end{figure*}

We propose CoGeo-GS, a multi-object removal framework built on concept-driven Gaussian representations. As shown in Fig.~\ref{fig:pipeline}, our framework integrates concept-aware semantic tagging, depth-guided geometric completion, and geometry-regularized appearance refinement. Together, these modules produce semantically coherent and geometrically stable 3D reconstructions across multiple views. 



\subsection{Background and Problem Definition}
\label{sec:prelim}
3D Gaussian Splatting (3DGS) represents a scene as a set of Gaussians $\mathcal{G} = \{g_i\}_{i=1}^{N}$ and renders images via differentiable $\alpha$-blending. Each Gaussian $g_i = (\boldsymbol{\mu}_i, \boldsymbol{\Sigma}_i, \alpha_i, \mathbf{c}_i)$ is defined by a 3D center $\boldsymbol{\mu}_i$, covariance $\boldsymbol{\Sigma}_i$, opacity $\alpha_i$, and SH-based color coefficients $\mathbf{c}_i$.
After projection onto the image plane, the color $C$ at a pixel is computed via $\alpha$‑blending of depth‑sorted Gaussians:
\begin{equation}
    C = \sum_{i \in \mathcal{N}} \mathbf{c}_i \alpha_i T_i, \quad
    T_i = \prod_{j=1}^{i-1} (1 - \alpha_j),
    \label{eq:alpha_blend}
\end{equation}
where $\mathcal{N}$ is the set of Gaussians overlapping the pixel and $T_i$ is the accumulated transmittance.

Given a reconstructed 3D Gaussian scene $\mathcal{G}$ and a set of concept masks that specify the objects to be removed, our task is to modify $\mathcal{G}$ such that the resulting scene remains geometrically complete and visually consistent. This requires solving three coupled challenges. First, all Gaussians associated with the target objects must be reliably identified and removed. Second, the geometry missing from the removed regions must be plausibly completed while preserving the surrounding background structure. Third, the final scene should support photorealistic rendering with appearance consistent across all views. Our goal is a semantically correct, geometrically stable, and visually coherent Gaussian field after removal.


\subsection{Concept-Aware Tagging for Multi-Object Removal}
\label{sec:concept}
To enable fine-grained object-level editing, we assign each Gaussian primitive a unique object identity that is consistent across views. Direct 2D-to-3D projection yields view-inconsistent labels, so we use offline text-driven preprocessing. For each target object, we provide a text prompt and use Grounding DINO to generate text-conditioned spatial prompts (e.g., bounding boxes), which guide SAM to extract the corresponding masks in each view. Since all views share the same set of prompts and indexing order, assigning label $q$ to pixels within the mask of the $q$-th prompt \textit{produces a cross-view consistent label map $O$}. This map satisfies $O(u)\in\{0,1,\ldots,Q\}$, where 0 denotes the background and $q\in\{1,\ldots,Q\}$ indexes the $Q$ target objects.

We further distill the 2D label map $O$ into the 3D Gaussian field by learning per-primitive feature vectors. Specifically, we assign a learnable feature vector $f_i\in\mathbb{R}^D$ to each Gaussian $i$. During rendering, we \textit{apply} Gaussian splatting to these features and perform front-to-back $\alpha$-compositing to obtain a per-pixel feature map $F(u)$:
\begin{equation}
    F(u)=\sum_{i\in\mathcal{N}_u}\alpha_i(u)\,T_i(u)\,f_i, T_i(u)=\prod_{j<i}\bigl(1-\alpha_j(u)\bigr),
    \label{eq:feature_rendering}
\end{equation}
where $u=(x,y)$ denotes pixel coordinates, $\mathcal{N}_u$ is the set of Gaussians contributing to pixel $u$ (sorted by depth), $\alpha_i(u)$ is the opacity contribution, and $T_i(u)$ is the accumulated transmittance. 
We apply a linear classifier $\Phi:\mathbb{R}^D\to\mathbb{R}^{Q+1}$ followed by the softmax function to obtain per-pixel object identity predictions:
\begin{equation}
    \bar{O}(u)=\mathrm{softmax}\bigl(\Phi(F(u))\bigr).
    \label{eq:pixel_pred}
\end{equation}
The predictions $\bar{O}(u)$ are aligned with $O(u)$ using a multi-class cross-entropy loss over the pixel set $P$:
\begin{equation}
L_{\mathrm{obj}}
= -\frac{1}{|P|}
\sum_{u \in P}
\mathbb{E}_{c \sim O(u)}\big[\log \bar{O}_c(u)\big].
\end{equation}

To propagate 3D supervision to occluded Gaussians near geometric boundaries, we enforce neighborhood consistency in 3D space. Let $x_i$ be the center of Gaussian $i$ and $\mathcal{K}(x_i)$ denote its $k$-nearest neighbors. We define the object-identity distribution of each Gaussian as:
\begin{equation}
    p_i=\mathrm{softmax}\bigl(\Phi(f_i)\bigr),
    \label{eq:gaussian_prob}
\end{equation}
where $\Phi$ shares weights with the 2D pixel classification branch. Let $\Omega$ denote the set of Gaussians sampled for computing $L_{\mathrm{space}}$. The spatial consistency loss is defined as:
\begin{equation}
    L_{\mathrm{space}}=\frac{1}{|\Omega|}\sum_{i\in\Omega}\frac{1}{k}\sum_{j\in\mathcal{K}(x_i)} \mathrm{KL}\bigl(p_i\,\|\,p_j\bigr).
    \label{eq:loss_space}
\end{equation}
The final distillation objective is utilized to optimize each identity in Gaussians:
\begin{equation}
    L_{\mathrm{Dis}}=L_{\mathrm{obj}}+\lambda\,L_{\text{space}}.
    \label{eq:loss_total}
\end{equation}
Through the synergy of 2D supervision and 3D regularization, each Gaussian gradually acquires a stable, cross-view consistent object identity, enabling subsequent object-level editing.

\subsection{Depth-Guided Geometric Completion}
\label{sec:depth}

Removing target Gaussians creates geometric holes. RGB inpainting with back-projection often breaks cross-view consistency, while diffusion-based depth completion without scale anchoring can drift in scale. We therefore propose a two-stage depth recovery pipeline that anchors global scale and then refines local geometric details.

\noindent\textbf{Stage 1: Scale Anchoring with Monocular Priors.}
Given the inpainted image $I_{\text{Inp}}^{(v)}$ and hole mask $M^{(v)}$ for view $v$, we apply the monocular depth model Depth Anything 3 to estimate relative depth $D_{\text{mono}}^{(v)}$. Since monocular predictions suffer from scale ambiguity and global shift, we align them to the current 3DGS geometry using reliable background regions.

Specifically, we estimate scale and shift parameters $(a^*, b^*)$ by least-squares fitting between the monocular prediction and the rendered depth $D_{\text{render}}^{(v)}$ over the background region $\Omega_{\text{bg}}^{(v)}$:
\begin{equation}
    (a^*, b^*) = \arg\min_{a,b} \sum_{u \in \Omega_{\text{bg}}^{(v)}} 
    \left( a D_{\text{mono}}^{(v)}(u) + b - D_{\text{render}}^{(v)}(u) \right)^2.
\end{equation}
The aligned depth is then obtained as $D_{\text{align}}^{(v)} = a^* D_{\text{mono}}^{(v)} + b^*$. To ensure smooth transitions at hole boundaries, we blend the aligned depth with the rendered depth using a softly dilated mask $M_{\text{soft}}^{(v)}$:
\begin{equation}
    D_{\text{anchor}}^{(v)} = M_{\text{soft}}^{(v)} \odot D_{\text{align}}^{(v)} 
    + (1 - M_{\text{soft}}^{(v)}) \odot D_{\text{render}}^{(v)}.
\end{equation}
This anchor depth provides a globally consistent geometric scaffold for subsequent refinement.

\noindent\textbf{Stage 2: Detail Refinement with Diffusion Priors.}
While $D_{\text{anchor}}^{(v)}$ ensures correct scale and coarse structure, monocular priors alone are insufficient to recover fine-grained geometry in heavily occluded regions. We therefore employ a pre-trained depth diffusion model (InFusion) as a plug-and-play prior to refine local details.

The diffusion process is conditioned on the inpainted RGB image, the anchor depth, and the hole mask by concatenating their latent representations:
\begin{equation}
    z_{\text{in}} = \mathrm{concat}\bigl( z_{\text{rgb}}, z_{d,t}, z_{\text{mask}}, z_{\text{mask\_d}} \bigr).
\end{equation}
To prevent geometric drift during sampling, we enforce a scale-anchored masked constraint at each diffusion step, explicitly fixing the background depth to the noised anchor depth:
\begin{equation}
    z_{d,t} \leftarrow (1 - z_{\text{mask}}) \odot z_{\text{anchor\_noised},t} 
    + z_{\text{mask}} \odot z_{d,t}.
\end{equation}
This constraint restricts the diffusion model to synthesize high-frequency geometry only within missing regions, while preserving global structure and scale consistency. The final completed depth $D_{\text{inpaint}}^{(v)}$ is obtained via reverse diffusion.

\subsection{Gaussian Fusion and Geometry-Regularized Refinement}
\label{sec:fusion}
After obtaining the inpainted depth $D_{\text{inpaint}}^{(v)}$, we back-project it into 3D space to generate a point cloud and convert it into a set
of Gaussians $\mathcal{G}_{\text{patch}}$. To mitigate redundant geometry and floater artifacts at the fusion interface, we implement an overlap-aware pruning strategy targeting the background Gaussian set $\mathcal{G}_{\text{bg}}$ near the hole boundary. Specifically, a background Gaussian $i \in \mathcal{G}_{\text{bg}}$ is pruned if it satisfies both of the following criteria:
\begin{itemize}[leftmargin=*,topsep=3pt,itemsep=2pt]
    \item \textbf{Proximity overlap:} It is spatially adjacent to the new patch: $\min_{k \in \mathcal{G}_{\text{patch}}} \|\mu_i - \mu_k\| < \delta_{\text{near}}$, where $\mu_i$ and $\mu_k$ denote the 3D center positions of Gaussians $i$ and $k$, respectively.
    \item \textbf{Local sparsity:} It is identified as a sparse outlier, i.e., the neighbor count within a radius $r$ falls below $n_{\text{min}}$.
\end{itemize}
The remaining background Gaussians are merged with $\mathcal{G}_{\text{patch}}$ to form the initial fused scene $\mathcal{G}_{\text{init}}$.

To enhance consistency and prevent structural drift, we adopt a two-stage refinement strategy with geometric regularization.
We first fix all geometric parameters and optimize only SH coefficients and opacity, allowing the newly inserted Gaussians to match surrounding appearance without altering the underlying structure. Using the inpainted image $I_{\text{Inp}}^{(v)}$ as reference, we minimize the photometric loss:
\begin{equation}
    L_{\text{ref}} = (1 - \lambda_{\text{ssim}})\,\|I_{\text{render}} - I_{\text{Inp}}^{(v)}\|_1
    + \lambda_{\text{ssim}}\bigl(1 - \text{SSIM}(I_{\text{render}}, I_{\text{Inp}}^{(v)})\bigr).
\end{equation}
Subsequently, we enable updates to geometric parameters to eliminate residual seams at the fusion boundary. To prevent unintended deformation of unedited regions, we constrain each geometric update $\Delta\theta$ within a trust region by norm clipping:
\begin{equation}
    \Delta\theta \leftarrow \Delta\theta \cdot \min\!\left(1, \frac{\varepsilon}{\|\Delta\theta\|_2}\right).
\end{equation}
This constraint is applied independently to each parameter tensor, enabling local seam correction while maintaining global geometric consistency.


\section{Experiments}

\subsection{Experimental Setup}

\noindent\textbf{Datasets.} We evaluate CoGeo-GS on the standard public benchmarks Mip-NeRF 360~\cite{barron2022mip} and SPIn-NeRF~\cite{mirzaei2023spin}. These datasets are selected to comprehensively validate the effectiveness of our method across both bounded and unbounded environments. To ensure fair benchmarking, all methods are evaluated using the provided camera intrinsics and initialized SfM point clouds from COLMAP~\cite{schonberger2016structure}.

\noindent\textbf{Baselines and Metrics.} We compare our method with four state-of-the-art 3D inpainting and editing methods: InFusion~\cite{liu2024infusion}, SPIn-NeRF~\cite{mirzaei2023spin}, GaussianEditor~\cite{chen2024gaussianeditor} and Gaussian Grouping~\cite{ye2024gaussian}. For fair comparison, we retrain all baseline models on the benchmarks using their official open-source implementations.  We employ PSNR, SSIM, LPIPS, and FID to evaluate visual quality. Crucially, to strictly assess the synthesis quality of the edited content, all metrics are computed exclusively on the masked regions. 

\noindent\textbf{Implementation Details.}
Experiments run on a single NVIDIA RTX 4090 GPU. 3DGS optimization uses 30k iterations at original resolution. In the concept-aware distillation stage, we set the object feature dimension to $D=16$ and apply spatial consistency regularization using $k=5$ nearest neighbors with weight $\lambda=0.1$. 
For geometric completion, monocular depth estimation is performed at $504{\times}504$ resolution to provide global scale anchors, followed by diffusion-based refinement at $768{\times}768$ resolution for detail recovery.
After fusion, the scene is fine-tuned for 150 iterations using an $\mathcal{L}_1$ and SSIM loss with $\lambda_{\text{ssim}}=0.2$. To ensure stability, geometric parameters are frozen during the initial optimization phase and subsequently updated under a trust-region constraint.

\begin{table*}[!t]
\centering
\caption{Quantitative comparison on multi-object and single-object removal tasks.
Higher PSNR and SSIM indicate better rendering fidelity, while lower LPIPS and FID
indicate better perceptual quality.}
\label{tab:main_benchmark}
\footnotesize
\begingroup
\setlength{\tabcolsep}{7pt}
\renewcommand{\arraystretch}{1.12}

\begin{tabularx}{\textwidth}{l*{4}{>{\centering\arraybackslash}X} >{\hspace{8pt}\centering\arraybackslash}X *{3}{>{\centering\arraybackslash}X}} 
\toprule
\multirow{2}{*}{\textbf{Method}} & \multicolumn{4}{c}{\textbf{Task 1: Multi-Object Removal}} & \multicolumn{4}{c}{\textbf{Task 2: Single-Object Removal}} \\
\cmidrule(lr){2-5} \cmidrule(lr){6-9} 
 & PSNR$\uparrow$ & SSIM$\uparrow$ & LPIPS$\downarrow$ & FID$\downarrow$ 
 & PSNR$\uparrow$ & SSIM$\uparrow$ & LPIPS$\downarrow$ & FID$\downarrow$ \\
\midrule
GaussianEditor~\cite{chen2024gaussianeditor} 
& 22.4 & 0.742 & 0.284 & 31.6 
& 24.1 & 0.781 & 0.247 & 28.4 \\

InFusion~\cite{liu2024infusion} 
& 24.8 & 0.803 & 0.213 & 24.9 
& 26.5 & 0.831 & 0.186 & 21.7 \\

Gaussian Grouping~\cite{ye2024gaussian} 
& 25.6 & 0.821 & 0.198 & 22.8 
& 27.3 & 0.847 & 0.172 & 19.6 \\

SPIn-NeRF~\cite{mirzaei2023spin} 
& 24.9 & 0.812 & 0.221 & 23.7 
& 26.8 & 0.842 & 0.189 & 20.9 \\

\rowcolor{blue!5}
\textbf{CoGeo-GS (Ours)} 
& \textbf{28.9} & \textbf{0.882} & \textbf{0.086} & \textbf{11.4} 
& \textbf{30.7} & \textbf{0.903} & \textbf{0.072} & \textbf{9.8} \\

\midrule
\bottomrule
\end{tabularx}
\endgroup
\vspace{-0.3cm}
\end{table*}

\begin{figure*}[!t]
  \centering
  \includegraphics[width=\linewidth,angle=0]{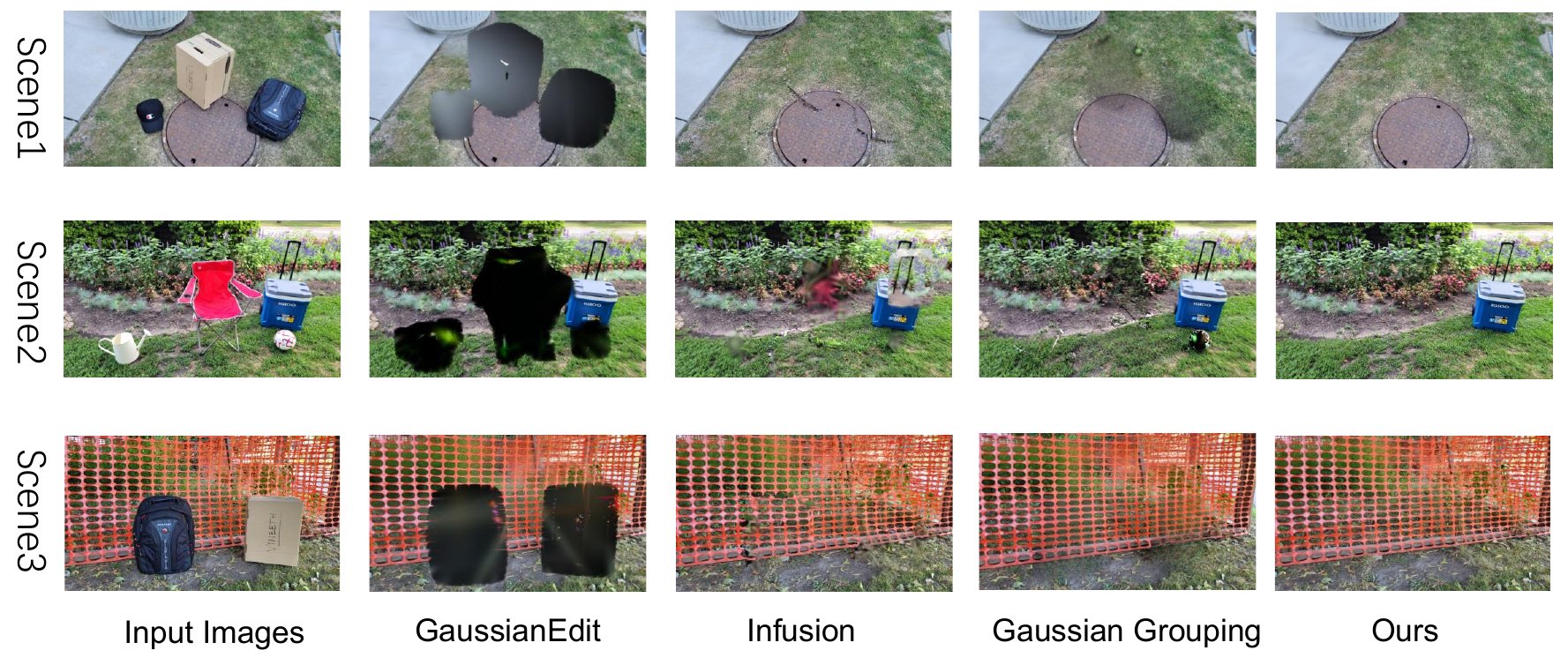}
    \caption{Qualitative comparison on multi-object removal. We compare CoGeo-GS with GaussianEditor, InFusion and Gaussian Grouping in a consistent view.}
  \label{fig:comparison}
  \vspace{-0.4cm}
\end{figure*}

\subsection{Comparison Results}
We evaluate our method against state-of-the-art baselines on both multi-object and single-object removal tasks. 

\noindent\textbf{Multi-object Removal Results.} 
As shown in Fig.~\ref{fig:comparison}, GaussianEditor~\cite{chen2024gaussianeditor} improves appearance but often introduces obviously black holes in removing multiple objects. InFusion~\cite{liu2024infusion} tends to produce similar results and inconsistent completion occurs near boundaries, causing floating artifacts due to scale mismatches. Gaussian Grouping~\cite{ye2024gaussian} removes multiple objects but suffers from floating artifacts and inconsistent appearance between background and inpainting regions. In contrast, our CoGeo-GS utilizes depth-guided completion to maintain structural integrity across views. Quantitative results are demonstrated in Task 1 of Tab.~\ref{tab:main_benchmark}. For multi-object removal, CoGeo-GS improves PSNR and SSIM over the strongest baseline, while reducing LPIPS and FID by 56.6\% and 50.0\%, respectively. Our method effectively handles complex occlusions significantly better than existing approaches.

\begin{figure}[!t]
  \centering
  \includegraphics[width=\linewidth,angle=0]{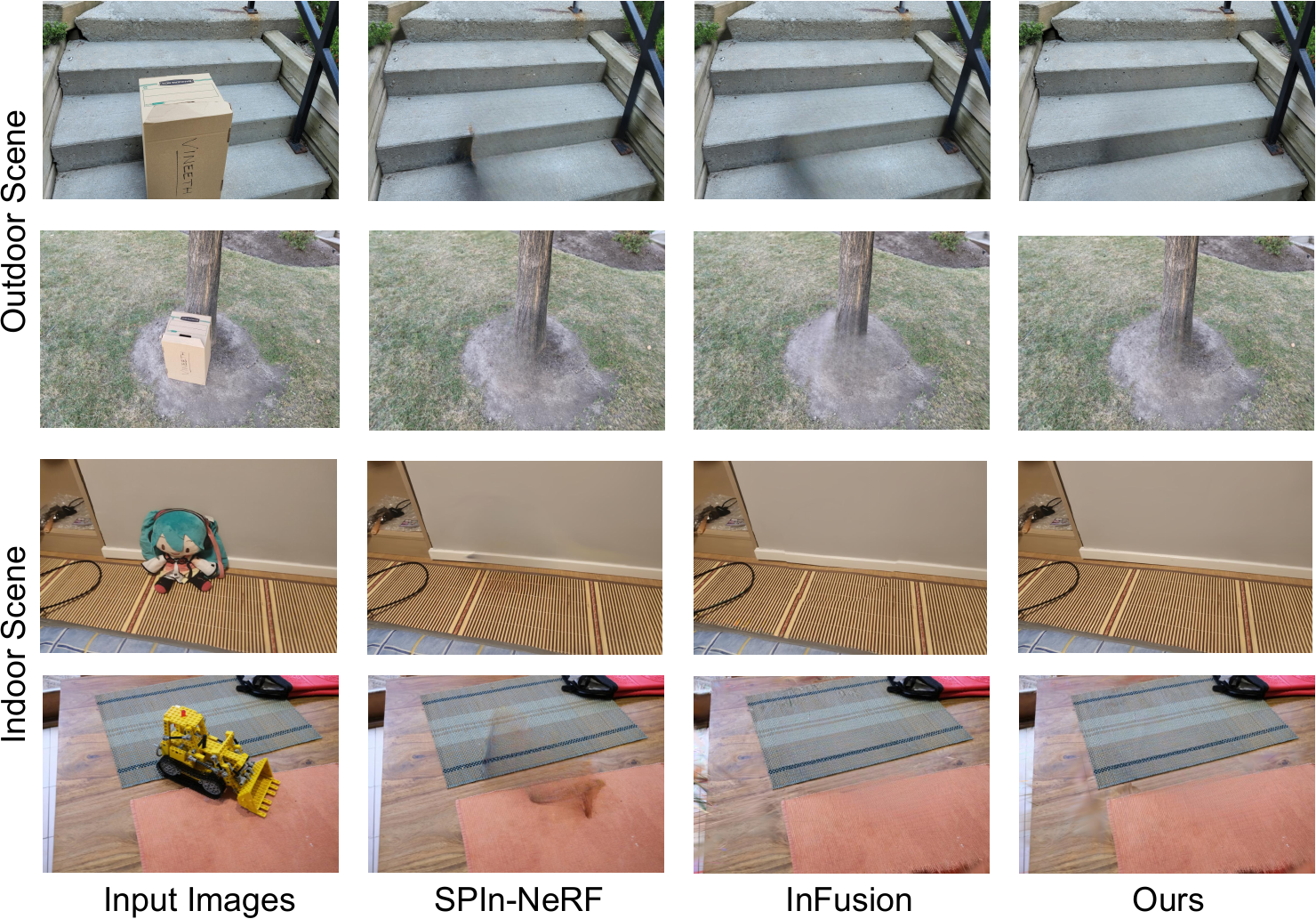}
    \caption{Qualitative comparison on single-object removal. We compare CoGeo-GS with SPIn-NeRF and InFusion in a consistent view.}
  \label{fig:comparison-single}
  \vspace{-0.2cm}
\end{figure}

\noindent\textbf{Single-object Removal Results.} Fig.~\ref{fig:comparison-single} presents the comparison for single-object removal. SPIn-NeRF~\cite{mirzaei2023spin} maintains 3D consistency but requires slow per-scene optimization and often produce blurry textures in large masked regions. {InFusion} exhibits better texture but lacks geometric stability. Our method preserves clear, high-frequency details consistent with the surrounding texture. This indicates that our concept-aware distillation ensures precise object removal without degrading the visual quality of the remaining scene.
As reported in Task 2 of Tab.~\ref{tab:main_benchmark}, CoGeo-GS outperforms baselines on the SPIn-NeRF dataset, achieving PSNR, SSIM, LPIPS and FID improvement. These gains demonstrate the superiority of CoGeo-GS in improving both geometric fidelity and perceptual quality.

\begin{table}[!t]
\centering
\footnotesize
\caption{Ablation study on the multi-object removal benchmark.}
\label{tab:ablation}
\setlength{\tabcolsep}{5pt}
\begin{tabular}{lcccc}
\toprule
Method & PSNR$\uparrow$ & SSIM$\uparrow$ & LPIPS$\downarrow$ & FID$\downarrow$ \\
\midrule
w/o concept-aware masking  & 26.85 & 0.851 & 0.112 & 16.90 \\
w/o depth guidance         & 25.97 & 0.832 & 0.128 & 18.40 \\
\midrule
Full model (Ours)          & \textbf{28.90} & \textbf{0.882} & \textbf{0.086} & \textbf{11.40} \\
\bottomrule
\end{tabular}
\vspace{-0.3cm}
\end{table}

\subsection{Ablation Study}
\begin{figure}[!t]
  \centering
  \includegraphics[width=\linewidth,angle=0]{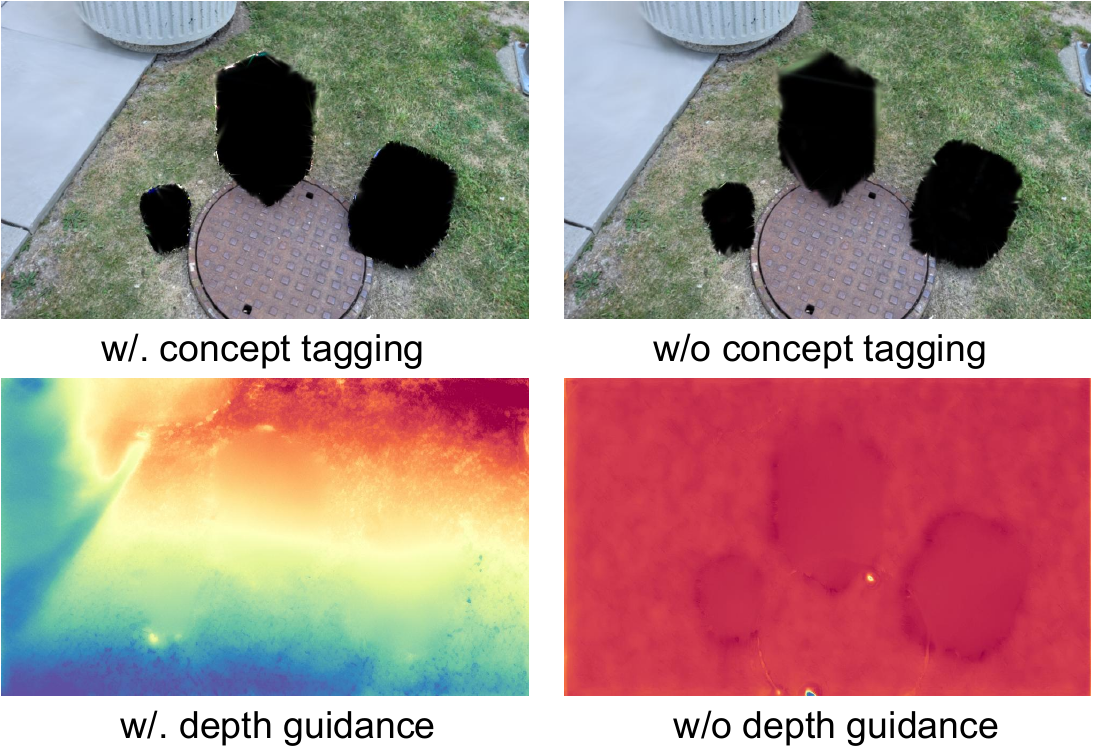}
  \caption{Ablation study. 
  Without concept-aware masking, residual artifacts persist in the edited regions. In the absence of depth guidance, the predicted depth map leads to blurred shadows and degraded structure in occluded areas.
  }
  \label{fig:ablation}
  \vspace{-0.3cm}
\end{figure}

We evaluate the effect of two core components: concept-aware 3D masking and depth-guided geometric completion. 
We replace the proposed 3D concept distillation with a naive 2D projection baseline, where a Gaussian is removed if its projected center lies inside the object mask in any view. As shown in the first row of Tab.~\ref{tab:ablation}, this leads to noticeable performance degradation, particularly in PSNR. The degradation arises from view-dependent visibility and partial occlusions, which cannot be reliably handled by 2D projections.
To assess the role of geometric anchoring, we remove the monocular depth prior and perform diffusion-based depth completion without scale guidance. As reported in the second row of Tab.~\ref{tab:ablation}, this variant suffers from clear drops in PSNR and FID, accompanied by floating artifacts and geometric misalignment in Fig.~\ref{fig:ablation}. Without depth anchoring, the reconstructed Gaussians fail to maintain consistent surface geometry.

\FloatBarrier

\section{Conclusion}
\vspace{-0.1cm}
We introduced CoGeo-GS, a concept-driven framework tailored for multi-object removal in 3D scenes. By unifying concept-aware 3D Gaussian selection, depth-guided geometric completion, and geometry-regularized appearance refinement within a single optimization pipeline, CoGeo-GS reliably reconstructs occluded background geometry while preserving cross-view consistency. Extensive experiments on multi-object removal demonstrate that our method achieves improved geometric fidelity and perceptual quality, consistently outperforming prior Gaussian-based editing approaches. We believe CoGeo-GS provides promising directions toward interactive 3D content manipulation. 
\section*{Acknowledgement}
This project was supported by the National Natural Science Foundation of China under Grant No.~62472415. 


\bibliographystyle{IEEEbib}
\bibliography{CoGeo-GS}

@String(CVPR = {Proc. IEEE/CVF Conf. Comput. Vis. Pattern Recognit.})

@String(ICCV = {Proc. IEEE/CVF Int. Conf. Comput. Vis.})

@String(ECCV = {Proc. Eur. Conf. Comput. Vis.})

@String(NIPS = {Adv. Neural Inf. Process. Syst.})

@String(TOG= {ACM Trans. Graph.})

@String(TIP = {IEEE Trans. Image Process.})

@String(WACV = {Proc. IEEE/CVF Winter Conf. Appl. Comput. Vis.})

@String(VR = {Vis. Res.})

@article{yang2025drivemoe,
      title={DriveMoE: Mixture-of-experts for vision-language-action model in end-to-end autonomous driving},
      author={Yang, Zhenjie and Chai, Yilin and Jia, Xiaosong and Li, Qifeng and Shao, Yuqian and Zhu, Xuekai and Su, Haisheng and Yan, Junchi},
      journal={arXiv preprint arXiv:2505.16278},
      year={2025}
}

@inproceedings{huang2023iddr,
  title={Iddr-ngp: Incorporating detectors for distractors removal with instant neural radiance field},
  author={Huang, Xianliang and Gou, Jiajie and Chen, Shuhang and Zhong, Zhizhou and Guan, Jihong and Zhou, Shuigeng},
  booktitle={Proceedings of the 31st ACM International Conference on Multimedia},
  pages={1343--1351},
  year={2023}
}

@inproceedings{huang2024hi,
  title={Hi-nerf: Hybridizing 2d inpainting with neural radiance fields for 3d scene inpainting},
  author={Huang, Xianliang and Chen, Shuhang and Zhong, Zhizhou and Gou, Jiajie and Guan, Jihong and Zhou, Shuigeng},
  booktitle={Proceedings of the Asian Conference on Computer Vision},
  pages={2855--2871},
  year={2024}
}

@article{huang2026nerf,
  title={NeRF-MIR: Toward High-Quality Restoration of Masked Images With Neural Radiance Fields},
  author={Huang, Xianliang and Zhong, Zhizhou and Chen, Shuhang and Xu, Yi and Guan, Jihong and Zhou, Shuigeng},
  journal={IEEE Transactions on Neural Networks and Learning Systems},
  year={2026},
  publisher={IEEE}
}

@inproceedings{mirzaei2023spin,
  title={Spin-nerf: Multiview segmentation and perceptual inpainting with neural radiance fields},
  author={Mirzaei, Ashkan and Aumentado-Armstrong, Tristan and Derpanis, Konstantinos G and Kelly, Jonathan and Brubaker, Marcus A and Gilitschenski, Igor and Levinshtein, Alex},
  booktitle=CVPR,
  pages={20669--20679},
  year={2023}
}

@inproceedings{schonberger2016structure,
  title={Structure-from-motion revisited},
  author={Schonberger, Johannes L and Frahm, Jan-Michael},
  booktitle={Proc. IEEE Conf. Comput. Vis. Pattern Recognit.},
  pages={4104--4113},
  year={2016}
}

@article{huang2026semantic,
  title={Semantic-Guided Progressive Object Removal with Gaussian Splatting},
  author={Huang, Xianliang and Xiao, Chen and Ni, Yuanxiang and Liu, Guanming and Liu, Mingkai and Fan, Dikai and Liu, Xiao and Zhang, Hao},
  journal={arXiv preprint arXiv:2607.04144},
  year={2026}
}

@article{ho2020denoising,
  title={Denoising diffusion probabilistic models},
  author={Ho, Jonathan and Jain, Ajay and Abbeel, Pieter},
  journal=NIPS,
  volume={33},
  pages={6840--6851},
  year={2020}
}

@inproceedings{ye2024gaussian,
  title={Gaussian grouping: Segment and edit anything in 3d scenes},
  author={Ye, Mingqiao and Danelljan, Martin and Yu, Fisher and Ke, Lei},
  booktitle=ECCV,
  pages={162--179},
  year={2024},
  organization={Springer}
}

@article{ruvzic2014context,
  title={Context-aware patch-based image inpainting using Markov random field modeling},
  author={Ru{\v{z}}i{\'c}, Tijana and Pi{\v{z}}urica, Aleksandra},
  journal=TIP,
  volume={24},
  number={1},
  pages={444--456},
  year={2014},
  publisher={IEEE}
}

@article{song2020score,
  title={Score-based generative modeling through stochastic differential equations},
  author={Song, Yang and Sohl-Dickstein, Jascha and Kingma, Diederik P and Kumar, Abhishek and Ermon, Stefano and Poole, Ben},
  journal={arXiv preprint arXiv:2011.13456},
  year={2020}
}

@inproceedings{xie2023smartbrush,
  title={Smartbrush: Text and shape guided object inpainting with diffusion model},
  author={Xie, Shaoan and Zhang, Zhifei and Lin, Zhe and Hinz, Tobias and Zhang, Kun},
  booktitle=CVPR,
  pages={22428--22437},
  year={2023}
}

@article{prabhu2023inpaint3d,
  title={Inpaint3d: 3d scene content generation using 2d inpainting diffusion},
  author={Prabhu, Kira and Wu, Jane and Tsai, Lynn and Hedman, Peter and Goldman, Dan B and Poole, Ben and Broxton, Michael},
  journal={arXiv preprint arXiv:2312.03869},
  year={2023}
}

@inproceedings{barron2022mip,
  title={Mip-nerf 360: Unbounded anti-aliased neural radiance fields},
  author={Barron, Jonathan T and Mildenhall, Ben and Verbin, Dor and Srinivasan, Pratul P and Hedman, Peter},
  booktitle=CVPR,
  pages={5470--5479},
  year={2022}
}

@inproceedings{suvorov2022resolution,
  title={Resolution-robust large mask inpainting with fourier convolutions},
  author={Suvorov, Roman and Logacheva, Elizaveta and Mashikhin, Anton and Remizova, Anastasia and Ashukha, Arsenii and Silvestrov, Aleksei and Kong, Naejin and Goka, Harshith and Park, Kiwoong and Lempitsky, Victor},
  booktitle=WACV,
  pages={2149--2159},
  year={2022}
}

@article{criminisi2004region,
  title={Region filling and object removal by exemplar-based image inpainting},
  author={Criminisi, Antonio and P{\'e}rez, Patrick and Toyama, Kentaro},
  journal=TIP,
  volume={13},
  number={9},
  pages={1200--1212},
  year={2004},
  publisher={IEEE}
}

@inproceedings{zeng2021cr,
  title={Cr-fill: Generative image inpainting with auxiliary contextual reconstruction},
  author={Zeng, Yu and Lin, Zhe and Lu, Huchuan and Patel, Vishal M},
  booktitle=ICCV,
  pages={14164--14173},
  year={2021}
}

@inproceedings{lugmayr2022repaint,
  title={Repaint: Inpainting using denoising diffusion probabilistic models},
  author={Lugmayr, Andreas and Danelljan, Martin and Romero, Andres and Yu, Fisher and Timofte, Radu and Van Gool, Luc},
  booktitle=CVPR,
  pages={11461--11471},
  year={2022}
}

@article{kerbl20233d,
  title={3D Gaussian Splatting for Real-Time Radiance Field Rendering.},
  author={Kerbl, Bernhard and Kopanas, Georgios and Leimk{\"u}hler, Thomas and Drettakis, George},
  journal=TOG,
  volume={42},
  number={4},
  pages={139--1},
  year={2023}
}

@inproceedings{chen2024gaussianeditor,
  title={Gaussianeditor: Swift and controllable 3d editing with gaussian splatting},
  author={Chen, Yiwen and Chen, Zilong and Zhang, Chi and Wang, Feng and Yang, Xiaofeng and Wang, Yikai and Cai, Zhongang and Yang, Lei and Liu, Huaping and Lin, Guosheng},
  booktitle=CVPR,
  pages={21476--21485},
  year={2024}
}

@article{mao2024live,
  title={LIVE-GS: LLM Powers Interactive VR by Enhancing Gaussian Splatting},
  author={Mao, Haotian and Xu, Zhuoxiong and Wei, Siyue and Quan, Yule and Deng, Nianchen and Yang, Xubo},
  journal={arXiv preprint arXiv:2412.09176},
  year={2024}
}

@article{zhu20243d,
  title={3D Gaussian Splatting in Robotics: A Survey},
  author={Zhu, Siting and Wang, Guangming and Kong, Dezhi and Wang, Hesheng},
  journal={arXiv preprint arXiv:2410.12262},
  year={2024}
}

@inproceedings{wang2024learning,
  title={Learning 3D Geometry and Feature Consistent Gaussian Splatting for Object Removal},
  author={Wang, Yuxin and Wu, Qianyi and Zhang, Guofeng and Xu, Dan},
  booktitle=ECCV,
  pages={1--17},
  year={2024},
  organization={Springer}
}

@article{liu2024infusion,
  title={Infusion: Inpainting 3d gaussians via learning depth completion from diffusion prior},
  author={Liu, Zhiheng and Ouyang, Hao and Wang, Qiuyu and Cheng, Ka Leong and Xiao, Jie and Zhu, Kai and Xue, Nan and Liu, Yu and Shen, Yujun and Cao, Yang},
  journal={arXiv preprint arXiv:2404.11613},
  year={2024}
}

\end{document}